\documentclass[runningheads,anonymous]{llncs}

\usepackage{graphicx}
\usepackage{multirow}
\usepackage{multicol}
\usepackage{makecell}
\usepackage{graphicx}
\usepackage{mathrsfs}
\usepackage{hyperref}       
\usepackage{url}            
\usepackage{booktabs}       
\usepackage{amsfonts}       
\usepackage{nicefrac}       
\usepackage{microtype}      
\usepackage{xcolor}         
\usepackage{lipsum}
\usepackage{tabularx}
\begin{document}
\title{SL: Stable Learning in Source-Free Domain Adaption for Medical Image Segmentation}


\author{Yixin Chen, Yan Wang}

\institute{Peking University, Beihang University}

%
\maketitle              
\begin{abstract}
Deep learning techniques for medical image analysis usually suffer from the domain shift between source and target data. Most existing works focus on unsupervised domain adaptation (UDA). However, in practical applications, privacy issues are much more severe. For example, the data of different hospitals have domain shifts due to equipment problems, and data of the two domains cannot be available simultaneously because of privacy. In this challenge defined as Source-Free UDA, the previous UDA medical methods are limited. Although a variety of medical source-free unsupervised domain adaption (MSFUDA) methods have been proposed, we found they fall into an over-fitting dilemma called “longer training, worse performance.” Therefore, we propose the \textbf{Stable Learning (SL)} strategy to address the dilemma. ST is a scalable method and can be integrated with other research, which consists of Weight Consolidation and Entropy Increase. First, we apply Weight Consolidation to retain domain-invariant knowledge and then we design Entropy Increase to avoid over-learning. Comparative experiments prove the effectiveness of SL. We also have done extensive ablation experiments. Besides, We will release codes including a variety of MSFUDA methods.
\keywords{Source Free Unsupervised Domain Adaptation  \and Training Stability.}
\end{abstract}
\section{Introduction}


In recent years, deep convolutional neural networks (DCNNs) have emerged as highly effective tools \cite{Deeplabv3,Unet,he2016deep,wang2020deep}, with particular success in medical image analysis\cite{Reiss_2021_CVPR,He_2021_CVPR,Chang_2020_CVPR}.  However, deep networks are data-driven, and they usually degrade due to the distribution differences between the training and the test data. When the test data in the target domain has the domain shift from the training data in the source domain, the well-trained model will drop in performance. This is a common challenge in medical image processing, such as data from different hospitals and data in cross-modalities. Further, deep learning methods rely on high-cost manual annotations, making it more challenging to apply DCNNs generally. To address this issue, many researchers turn to designing unsupervised domain adaptation methods by transferring knowledge from the labeled source domain to the unlabeled target domain \cite{SIFA,Cycada,dou2018unsupervised,SIFA_v2}. As a result, unsupervised domain adaptation tasks have gotten great attention and inspired various methods\cite{zhang2019category,zou2018unsupervised,mei2020instance,Cycada}. However, most of them still rely on the source data. Data privacy and transmission issues in realistic application scenarios are worthy of attention. Under this consideration, UDA tasks evolve without access to source data. This more challenging variant is known as Source-Free Unsupervised Domain Adaptation (SFUDA),

In SFUDA, we can only access a well-trained model from the source data. The previous works are roughly divided into two perspectives: GAN-based methods \cite{yang2020unsupervised,Cycada,SIFA}, and non-GAN-based methods \cite{SFDAKim,liang2020we}. Adversarial training has some clear limitations: the training process is unstable and has high complexity, which always causes uncertainties, and clinicians are often unwilling to believe in synthetic data. Non-GAN-based methods require a self-training approach to adapt the source model to the target domain \cite{DPL,mm2021domain,OSUDA}. Most of the previous work by self-training has solved the domain shift between the source domain and the target domain without access to the source data and target labels. However, there are still issues caused by the unstable self-training process. The experimental results show that the existing SFUDA methods eventually suffer a dilemma "longer training, worse performance." As an illustration, \cite{DPL} reports they trained 2 epochs in the self-training stage. Strictly assuming that the data of the target domain is not labeled, it is difficult to determine the number of self-trained epochs. The performance of existing SFUDA methods \cite{DPL,OSUDA,mm2021domain,DAS} drops sharply when the number of epochs increases.

In this paper, we analyze the causes of the dilemma “longer training, worse performance” and propose the Stable Learning (SL) framework to stabilize the self-training process, which comprises Weight Consolidation and Entropy Increase. 

\textbf{Weight consolidation.} the data sets for domain adaption tasks can be divided into domain-specific and domain-invariant. The domain-invariant knowledge learned by the model can be used for both source domain and target domain. In contrast, domain-specific knowledge leads to the performance degradation of the source model in the target domain. Therefore, the self-training process on the target domain aims to remove the source-domain-specific knowledge from the source model and keep the domain-invariant knowledge as much as possible. Neural networks encode knowledge in the form of trainable parameters. As a result, keeping the domain-invariant knowledge is equivalent to not updating all parameters in the model. Hence, we propose Weight Consolidation to promote the model to update some parameters selectively.

\textbf{Entropy Increase.} The cause of “longer training, worse performance” can be described as over-learning of the samples. Therefore we devised a strategy to reduce the impact of the samples that the model has learned and thus avoid over-learning. In the section \ref{sec:EI}, we prove that the idea of this is equivalent to the entropy increase.
To summarize, we introduce a Stable Learning (SL) strategy for source-free domain adaptation in medical semantic segmentation, which addresses the over-fitting problem in existing SFUDA methods caused by the lack of labeled target-domain data. The SL strategy comprises of two components: Weight Consolidation and Entropy Increase, and can be integrated with various existing medical SFUDA methods and We have demonstrated the effectiveness of our approach in four SFUDA tasks using two public datasets.

\section{Related Works}

\subsection{Source-free Unsupervised Domain Adaptation}

Considering the privacy and transmission issues in realistic application scenarios, source-free unsupervised domain adaptation (SFUDA) has attracted more attention recently \cite{liang2020we,mm2021domain}. Different from general unsupervised domain adaptation, source-free means providing a well-trained model without access to the source data during the adaption. The solutions to source-free problems often consist of two perspectives. First, compare the source-domain features and target-domain features to achieve the alignment. Second, apply self-training process to generate pseudo labels and perform label training. Some methods contain only the former \cite{OSUDA}, some methods may contain only the latter \cite{SFDAKim,mm2021domain,DPL}, and the others contain both perspectives \cite{liang2020we,DAS}. In addition to self-training, another solution of SFUDA is to generate and restore the source domain samples and align the features in adversarial manners \cite{kurmi2021domain,xia2021adaptive}. Because our proposed method is to solve the problem of instability of the self-training process, we do not describe the generative model too much.


\textbf{SFUDA Segmentation}. \cite{mm2021domain} attempts to create pseudo labels of target data for the segmentation task. They proposed a Label-Denoising framework (LD), containing positive learning and negative learning. Positive learning uses an intra-class threshold, which aims to solve the class imbalance issue. The pseudo label is used to train the source model supervised. Similarly, \cite{DPL} proposed Denoised Pseudo-Labeling (DPL) to create low-noise pseudo labels. They use the prototype strategy to calibrate the noise in the pseudo label, which is the same with \cite{SFDAKim}. Instead of using the idea of generating pseudo-labels, OS \cite{OSUDA} uses the adaption of the batch normalization layer to achieve domain-wise alignment. Low-order batch statistics, such as mean and variance, are domain-specific and high-order batch parameters like $\gamma$ and $\beta$ are domain-invariant. DAS \cite{DAS} freezes classifier parameters to minimize self-entropy, which attempts to align the source-domain feature and target-domain feature. Then, they create aligned source model to create pseudo label, using intra-class threshold \cite{mm2021domain}. 

We compare with LD \cite{mm2021domain}, DPL\cite{DPL} and OS \cite{OSUDA} in our experiments. We believe that the DAS \cite{DAS} method is basically the same as the LD \cite{mm2021domain} method and can be seen as an incremental experiment of LD, so DAS is not used as a separate comparative method.
\subsection{Self-training}
Self-training is widely used in semi-supervised learning and the key idea is to train the current model by generated pseudo labels from the previous models \cite{tarvainen2017mean,lee2013pseudo}. This is exactly in line with unsupervised domain adaptation's task settings so that self-training is one of the common solutions for UDA and SFUDA \cite{DAS,DPL,mm2021domain,pan2020unsupervised,zou2018unsupervised}. Due to the introduction of unlabeled data, the researchers need to improve the quality of predicted pseudo labels as much as possible \cite{tarvainen2017mean,yang2019training}. In the DA tasks, the generation of pseudo labels encounters greater challenges owing to the domain shift. Therefore, the previous work spent lots of efforts on how to generate reliable pseudo labels, such as the aforementioned work \cite{mm2021domain,DPL}. However, only a few works focus on the issue of stability during the training process. For example, \cite{ATSO} addresses the lazy mimicking, which describes the phenomenon of that student model learning from the pseudo labels arrives at a plateau during self-training. The authors propose the asynchronous teacher-student optimization algorithm to alleviate this issue and reports competitive performance. Similarly, we also address the stability of the training process.

\section{Methodology}
In this section, we mainly explain the two parts of the proposed Stable Learning, the core is Weight Consolidation (WC), and then use Entropy Increase (EI) to smooth the over-fitting interface and further stabilize the self-training process.

\subsection{Unstable Self-Training}

In source-free Unsupervised Domain Adaption (UDA) setting, we are given a model $f^s:X^s\rightarrow Y^s$ trained from unreachable source domain $D^s=(X^s,Y^s)$, and an target domain without annotation $D^t=(X^t)$. The goal of source-free UDA is to obtain an adapted model $f^{x\rightarrow y}$, which performs well on the target domain distribution. We consider the image segmentation, which is a multi-label segmentation problem with $X \subset \mathbb{R}^{H\times W\times 3}$. 

We found a common dilemma when using SOTA SFUDA methods, such as LD \cite{mm2021domain} and DPL \cite{DPL}. When using the target domain data to self-train the source model, the training process is unstable and the model falls into over-fitting easily. \cite{DPL} indicates that they trained the target model with 2 epochs. Other researches also prove that self-learning gets saturated after 2 epochs and the accuracy starts to drop in the 3rd epochs \cite{ATSO}. In our experiments, it is found that the performance of DPL and LD increases rapidly at epoch 1 or 2, and then decreases with the increase of epoch. However, this training process is trick because it needs a labeled target-domain validation data set as supervision. Without such supervision, these SFUDA approaches would be meaningless.

These methods contribute since the performance of the model can be improved, but they are flawed due to the improvement of performance is short-lived and unstable. In this paper, we propose SL to improve the stability of self-supervised training without affecting any performance.

\subsection{Weight Consolidation}

\begin{figure}[t]
  \centering

   \includegraphics[width=1\linewidth]{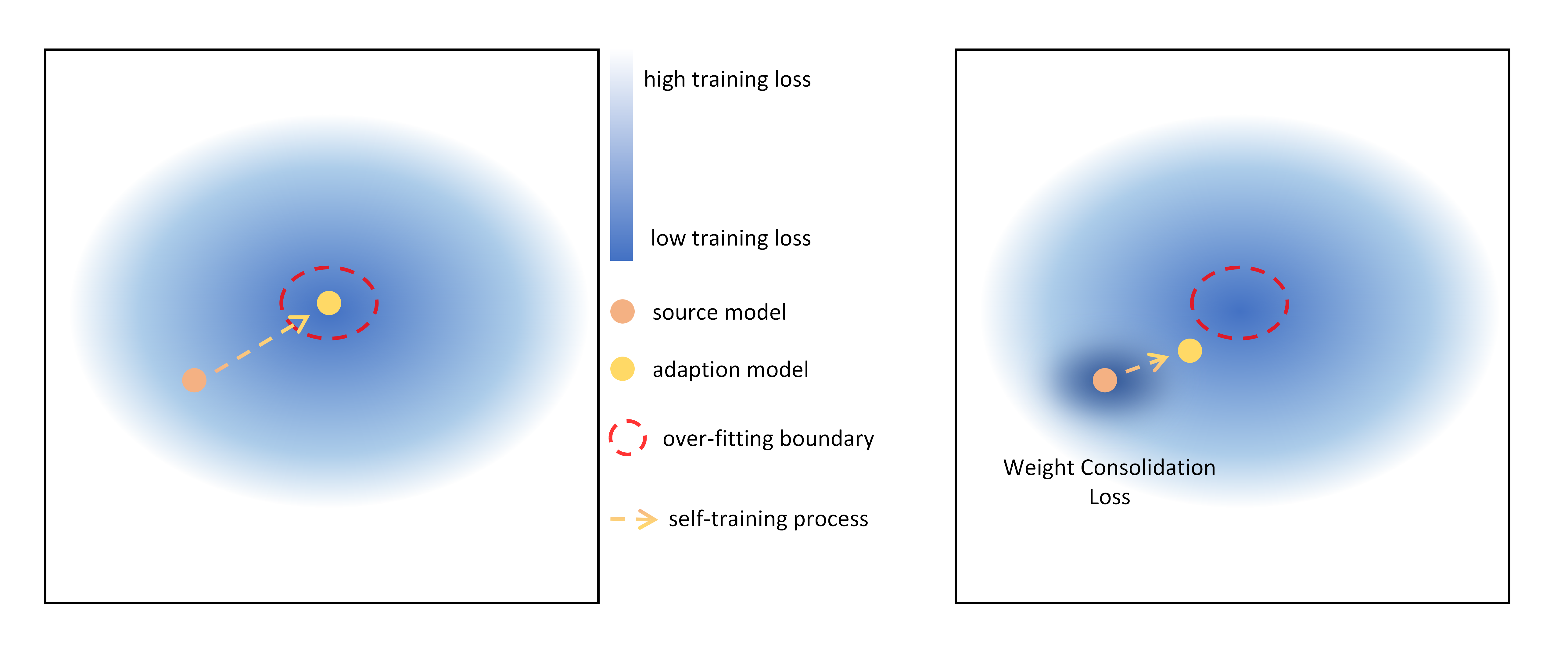}

   \caption{The comparison of domain adaptation with and without data Weight Consolidation loss. The subplot on the left shows traditional self-training process of SFUDA. The subplot on the right shows the self-training process with Weight Consolidation.}
   \label{fig:WC}
\end{figure}

\textbf{Motivation of weight consolidation}.  
Weight consolidation can be defined by adding weight penalty to the loss function:
\begin{equation}
    L'(\theta) = L(\theta) + \sum_{i}{|\theta_i-\theta^*_i|_1},
\end{equation}
where $\theta$ is the parameters set of model $f^{s\rightarrow t}$ and $\theta^*$ is the parameters set of the source model. $L(\theta)$ is the adaption loss, which might be different according to different MSFUDA methods. By imposing first norm penalties on the model parameters, only some parameters of the model can be updated. 

As shown in Figure \ref{fig:WC}, the left subplot shows the self-training process without weight consolidation. The parameters of the model will inevitably fall into the dilemma of over-fitting (Red circle) after enough iterations only under the guidance of adaption loss. Thus the cause of over-fitting can be simply summarized as the model completely forgetting the knowledge of source-domain-specific and domain-invariant. The right subplot shows the self-training process with weight consolidation. Imposing Weight consolidation helps allow the model to retain knowledge of source domains. Through adaption loss, the model can eliminate the negative impact of source-domain-specific knowledge and retain the knowledge of domain-invariant. Ultimately, the model avoids over-fitting outcomes and achieves stable training.

\textbf{No accuracy loss with weight consolidation}. The deep convolution network contains a large number of convolution layers and nonlinear activation function. Training process is to adjust the set of trainable parameters (weights, biases) of the convolution layers, to optimize performance. Many configurations of trainable parameters will result in the same performance \cite{hecht1992theory,sussmann1992uniqueness}. Therefore this overparameterization makes it possible to have a solution to make the model achieve the same performance under the constraint of weight consolidation. In addition, the experimental results also prove that weight consolidation can stabilize the MSFUDA methods without affecting the accuracy. The relevant experimental results can be found in Tables \ref{tab:abnorminal}.

\subsection{Entropy Increase}
\label{sec:EI}

The main purpose of weight consolidation is to prevent the model from forgetting domain-invariant knowledge. As shown in Figure \ref{fig:WC}, the weight consolidation also has the effect of preventing the model from over-fitting. However, WC alone does not completely avoid all over-fitting.

As mentioned earlier, "Many configurations of trainable parameters will result in the same performance \cite{hecht1992theory,sussmann1992uniqueness}." also contains this meaning: Many configurations of trainable parameters will result in the same over-fitting dilemma. Therefore this over-parameterization makes it possible to have a solution to make the model suffer the same over-fitting dilemma under the constraint of weight consolidation.

To address such a possibility, we propose Entropy Increase (EI) to avoid over-fitting during self-training. The cause of the over-fitting phenomenon can be described as over-learning of the samples. Therefore we devised a strategy to reduce the impact of learned samples on the model and thus avoid over-learning. Followed the setting of \cite{mm2021domain,DPL}, the adaption loss is cross entropy loss:

\begin{equation}
    L_a = -\hat{y}\log{p},
\end{equation}
where $L_a$ is the adaption loss, $\hat{y}$ is the pseudo label create by different methods and $p$ is the softmax prediction value of the model.

The learned sample is the sample that the model gives a high confidence, so the EI is to reduce the loss weight of the high confidence sample:

\begin{equation}
    L_a = -(\hat{y}-p)\log{p}=-\hat{y}\log{p} + p\log{p},
\end{equation}
Adaption loss using EI strategy can be broken down into two terms. The first is cross-entropy adaption loss and the second is self-entropy. Therefore, the essence of the EI strategy is to maximize the self-entropy of the prediction.

This conclusion is in some conflict with the results of previous studies: Entropy minimization has been shown effectiveness in semi-supervised learning \cite{NIPS2004_96f2b50b,springenberg2016unsupervised} and some MSFUDA methods also adopt entropy minimization loss as a part of them \cite{mm2021domain,OSUDA}. We believe that both self-entropy and cross-entropy are the directions of entropy minimization, which helps the convergence of the model, but promotes the model to fall into the dilemma of over-fitting, resulting in an unstable training process. In the semi-supervised task, a labeled validation set allows the model to select the moment of optimal performance during training. In the SFUDA mission, since there is no labeled target domain data, a stable training process becomes a necessary prerequisite. We conduct ablation experiments in the experimental part to compare the effects of entropy minimization and entropy maximization on model stability and performance.

\section{Experiments and discussions}
\begin{table*}
 \caption{Comparison of model training stability with SOTA unsupervised domain adaptation methods for Abdominal data set in Dice. The dice value is the average of the four categories of dice, and the four categories are Liver, R.kidney, L.kidney and Spleen.}
 \begin{center}
     
\scalebox{0.8}{
  \begin{tabular}{ccccccccc}
    \toprule
    Abdominal $CT \rightarrow MRI$ & \textbf{Epoch 1} & \textbf{Epoch 2} & \textbf{Epoch 3} & \textbf{Epoch 5} & \textbf{Epoch 10} & \textbf{Epoch 20} & \textbf{Epoch 50} &\textbf{Best Performance}\\
    \midrule
    Source Model&0.5167&-&-&-&-&-&-&-\\
    LD \cite{mm2021domain}  &0.6843&0.6904&0.6834&0.6803&0.6589&0.6146&0.5690&0.6904(epoch 1)\\
    fair LD   &0.6861&0.6843&0.6789&0.6843&0.6546&0.6114&0.5578&0.6861(epoch 1) \\
    DPL \cite{DPL}  &0.6673&0.6692&0.6615&0.6506&0.6128&0.5918&0.5375&0.6692(epoch 1) \\  
    OS \cite{OSUDA}  &0.6771&0.6728&0.6773&0.6762&0.6680&0.6796&0.6765&0.6846(epoch 25) \\
    fairLD with SL(ours) &0.6802&0.6729&0.6890&0.6918&0.6844&0.6893&0.6821&0.7006(epoch 16)\\
    DPL with SL (ours) & 0.6768&0.6707&0.6734&0.6812&0.6762&0.6896&0.6866&0.6961(epoch 36)\\  
    OS  with SL (ours) & 0.6667&0.6704&0.6647&0.6734&0.6574&0.6649&0.6673&0.6867(epoch 17) \\
    \bottomrule
  \end{tabular}
}

\scalebox{0.8}{
  \begin{tabular}{ccccccccc}
    \toprule
    Abdominal $MRI \rightarrow CT$ & \textbf{Epoch 1} & \textbf{Epoch 2} & \textbf{Epoch 3} & \textbf{Epoch 5} & \textbf{Epoch 10} & \textbf{Epoch 20} & \textbf{Epoch 50}&\textbf{Best Performance} \\
    \midrule
    Source Model&0.6474&-&-&-&-&-&-&-\\
    LD \cite{mm2021domain}  &0.5374&0.3905&0.3100&0.2141&0.1845&0.1717&0.1624&0.5374(epoch 1) \\
    fair LD   &0.6936&0.7029&0.7051&0.7034&0.6846&0.6777&0.6532&0.7061(epoch 5) \\
    DPL \cite{DPL}  &0.6534&0.6356&0.6218&0.6364&0.6156&0.6076&0.6097&0.6534(epoch 1) \\  
    OS \cite{OSUDA}  &0.6460&0.6447&0.6371&0.6580&0.6220&0.6251&0.5835&0.6580(epoch 4) \\
    fairLD with SL(ours) & 0.6437&0.6489&0.6669&0.6766&0.7025&0.7068&0.7128&0.7188(epoch 47)\\
    DPL with SL (ours) &0.6484&0.6435&0.6377&0.6361&0.6484&0.6426&0.6403&0.6535 (epoch 17) \\  
    OS  with SL (ours) & 0.6397&0.6502&0.6376&0.6217&0.6429&0.6317&0.6304&0.6606(epoch 23) \\
    \bottomrule
  \end{tabular}
}
  \label{tab:abnorminal}
   \end{center}
\end{table*}

\subsection{Dataset and Metrics}
We evaluated our proposed method on an abdominal data set, which is used by other Domain Adaption researches \cite{cyclegan,SIFA,Cycada,SASAN}. We follow the configuration of this abdominal dataset, which is made up of two groups of data: $20$ MRI scans from the CHAOS challenge\cite{chaos} and $30$ CT scans from Multi-Atlas Labeling Beyond the Cranial Vault- Workshop and Challenge \cite{landman2015multi}. Each MRI scan is a 3D volume of $256 \times 256 \times L$ voxels, where L is the length of the long axis. Each CT scan is a $512 \times 512 \times L$ 3D volume. According to the setting of \cite{SIFA}, we divide the training set and the validation set according to the ratio of 4:1. The ground truth masks are annotated as liver, right kidney (R-Kid), left kidney (L-Kid) and spleen. We proceeded with two domain adaption tasks, from CT to MRI and from MRI to CT.

The Dice coefficient (Dice) is the basic-used metric in segmentation tasks so that we also employed it to evaluate the performance. Dice is the measurement of volume overlap between the predictions and the ground truth annotations in 3D.

\section{Implementation Details}
\subsection{Training Configuration}
The configuration of the source model follows papers \cite{mm2021domain,DPL} and use deeplabv3-resnet50 segmentation model. We employ the following image augmentation strategies: Blur, ShiftScaleRotate, RandomBrightnessContrast, RandomGridShuffle. The batch size is set to 4 and we adopt Adam optimizor with 3e-5 learning rate and 3e-5 weight decay.

\subsection{Fairness optimization}
We fount that the pseudo labels generated by LD method \cite{mm2021domain} depend on the data set. In our experiments, it is found that the pseudo labels generated by LD \cite{mm2021domain} in abdominal data contain much noise, which affects the performance of self-training. We believe that this impact is due to the poor adaptability between the data set and the method, not the problem of the method itself. Therefore, the comparison under this premise is unfair and ineffective. We extend and improve the algorithm of generating pseudo labels by LD \cite{mm2021domain} method, which is called Double Threshold Pseudo Label (DTPL).

\begin{figure}[t]
  \centering
   \includegraphics[width=1\linewidth]{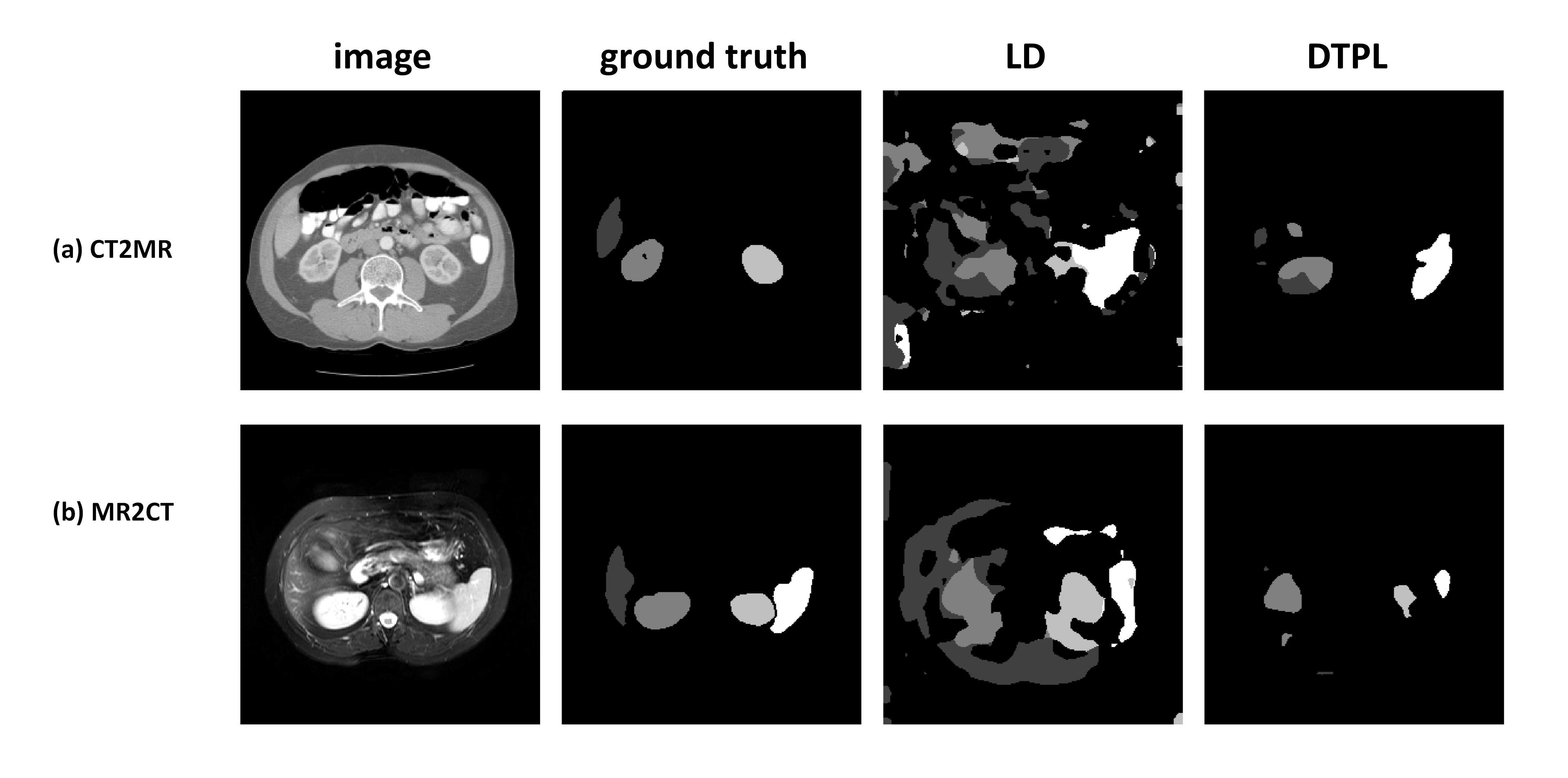}
   \caption{Comparison of pseudo labels in LD \cite{mm2021domain} and DTPL in abdominal data set. The four columns from left to right represent images, annotations, pseudo label generated by LD and pseudo label generated by DTPL.}
   \label{fig:DTPL}
\end{figure}

LD \cite{mm2021domain} use intra-class confidence to select the pixels as pseudo label, which can avoid the imbalanced selection and "winner-takes-all" dilemma (the model would be bias towards the majority classes and ignore the minority classes). They select the pixels with high intra-class confidence. The intra-class threshold is defined as :

\begin{equation}
    \delta^{(c)} = \tau_{\alpha}(p_t^{(c)}),
  \label{eq:2}
\end{equation}

where $p_t^{(c)}$ is prediction softmax values with respect to category $c$ and $\tau_{\alpha}$ means the top $\alpha (\%)$ value. Therefore, for each category, we select the labels whose softmax value is larger than threshold $\delta^{(c)}$. However, the quality of pseudo labels depends very much on the setting of $\alpha$. LD \cite{mm2021domain} use the same $\alpha$ between different categories, which is set as 0.3 but this does not work in any data sets. As shown in Figure \ref{fig:DTPL}, we found LD method create much noise. Different categories use the same $\alpha$ as their intra-class threshold, which causes each category in the pseudo-label has a similar area. We can avoid this problem by setting $\alpha$ separately for each category, but too many hyperparameters can bring a lot of complexity.

Experimentally, we found that the generated noise can be filtered out by a global threshold, so we use double thresholds (global threshold and intra-class threshold) to generate pseudo-labels. Double Threshold Pseudo Label can be defined as :

\begin{equation}
 \hat{y}^{h,w,c}=\left\{
\begin{array}{rcl}
1 && if  c=(argmax_c{p^{h,w,c}}) \cap (p^{h,w,c} > \delta^{(c)}) \\
&&\cap (p^{h,w,c} >\lambda)\\
0 && otherwise
\end{array}
\right.
\end{equation}
where $\hat{y}^{h,w,c}$ is the pixel-wise pseudo label. $(p^{h,w,c} > \delta^{(c)})$ represents intra-class threshold and $ (p^{h,w,c} >\lambda)$ means the global threshold. In our research, the $alpha$ is set to 0.3 and $\lambda$ is set to 0.2.

\subsection{Abdominal Comparison Experiments}
In Table \ref{tab:abnorminal}, we compare state-of-the-art approaches with or without Stable Learning strategy and show that approaches with ST are more stable. The table above shows the results of adaption task from CT to MRI modilaty and the table below shows the results of adaption task from MRI to CT. We compared snapshots of different methods at 1st, 2nd, 3rd, 5th, 10th, 20th and 50th epochs and found that the MSFUDA methods \cite{mm2021domain,DPL,OSUDA} without an SL policy would suffer a "longer training, worse performance" dilemma. For example, in task $MRI \rightarrow CT $, DPL performance at 50th epoch is 0.6097 but the best performance of DPL is 0.6580 at 1st epoch. Similarly, the performance of OS \cite{OSUDA} is 0.5835 at 50th epoch but its best performance is 0.6580 as 4th epoch. The methods with SL outperforms those without SL. The experimental data in Tabel \ref{tab:abnorminal} can fully prove this idea. In addition, the experimental results of two SFUDA tasks on the cardiac dataset are presented in the supplementary materials.

\subsection{Cardiac Comparison Experiments}

\begin{table*}
  \caption{Comparison of model training stability for Cardiac data set in Dice. The dice value is the average of the four categories of dice, and the four categories are ascending aorta, left atrium blood cavity, left ventricle blood cavity and myocardium of the left ventricle.}
  \begin{center}
 \scalebox{0.8}{
  \begin{tabular}{ccccccccc}
    \toprule
    Cardiac $CT \rightarrow MRI$ & \textbf{Epoch 1} & \textbf{Epoch 2} & \textbf{Epoch 3} & \textbf{Epoch 5} & \textbf{Epoch 10} & \textbf{Epoch 20} & \textbf{Epoch 50} &\textbf{Best Performance}\\
    \midrule
    Source Model& 0.3753 &-&-&-&-&-&-&-\\
    LD \cite{mm2021domain}  &0.4089&0.4080&0.4019&0.3874&0.3761&0.3678&0.3535&0.4089(epoch 0)\\
    fair LD   & 0.4155&0.4145&0.4202&0.4124&0.4022&0.3768&0.3661&0.4202(epoch 2)\\
    DPL \cite{DPL}  & 0.4036&0.4162&0.4096&0.4216&0.4125&0.4128&0.3961&0.4238(epoch 8)\\  
    OS \cite{OSUDA}  & 0.3974&0.4162&0.4140&0.4141&0.4107&0.4154&0.4186&0.4261(epoch 36)\\
    fairLD with SL(ours) &0.4073&0.4124&0.4150&0.4171&0.4165&0.4101&0.4136&0.4264(epoch 13)\\
    DPL with SL (ours) & 0.4109&0.4076&0.4154&0.4174&0.4151&0.4177&0.4213&0.4238(epoch 44)\\  
    OS  with SL (ours) &  0.4125&0.4194&0.4096&0.4154&0.4195&0.4195&0.4141&0.4202(epoch 24)\\
    \bottomrule
  \end{tabular}
}
\scalebox{0.8}{
  \begin{tabular}{ccccccccc}
    \toprule
    Cardiac $MRI \rightarrow CT$ & \textbf{Epoch 1} & \textbf{Epoch 2} & \textbf{Epoch 3} & \textbf{Epoch 5} & \textbf{Epoch 10} & \textbf{Epoch 20} & \textbf{Epoch 50}&\textbf{Best Performance} \\
    \midrule
    Source Model&0.4951 &-&-&-&-&-&-&-\\
    LD \cite{mm2021domain}  &0.5548&0.5479&0.5863&0.5280&0.5569&0.5557&0.5563&0.5990(epoch 8) \\
    fair LD &0.5840&0.5727&0.6060&0.6033&0.5985&0.5712&0.5829&0.6234(epoch 3) \\
    DPL \cite{DPL}  & 0.5447&0.5114&0.5393&0.4795&0.4641&0.5071&0.5049&0.5447(epoch 0)\\  
    OS \cite{OSUDA}  &0.4334&0.4398&0.4564&0.4479&0.4464&0.4609&0.4375&0.4609(epoch 19)\\
    fairLD with SL(ours) &0.5772&0.5816&0.5934&0.6213&0.6211&0.6119&0.6223&0.6289(epoch 28) \\
    DPL with SL (ours) &0.5211&0.5288&0.5340&0.5418&0.5591&0.5509&0.5456&0.5608(epoch 14) \\  
    OS  with SL (ours) & 0.4834&0.4998&0.5012&0.4991&0.4867&0.4950&0.4912&0.5083(epoch 22)  \\
    \bottomrule
  \end{tabular}
}
  \label{tab:cardiac}
  \end{center}
\end{table*}

In Table \ref{tab:cardiac}, we compare state-of-the-art methods with or without Stable Learning strategy. The result proves the efficiency of ST. The experimental design is consistent with that on the abdominal data set. In 50th epoch, the performance of fairLD and DPL are 0.3661 and 0.3961 respectively, which has a huge drop off from the best performance (0.4202 and 0.4238). However, under the help of stable learning strategy, the performance of fairLD and DPL are 0.4136 and 0.4213 in the 50th epoch, which is consistent with the best performance (0.4264 and 0.4238 respectively).

\subsection{Ablation Study}
The results of ablation experiments are shown in Table \ref{tab:ablation}. The ablation experiment consists of four source-free domain adaption tasks: two tasks on abdominal data set and the other two on the cardiac data set. In this section, we analyze three aspects: first, whether self-entropy minimization plays a positive role in SFUDA task; second, whether Stable Learning stabilizes the self-training; third, the impact of entropy increase in stable learning. 

\begin{table*}
 \caption{The results of the ablation experiments. The table above shows the experimental results on abdominal data set and the table below is about cardiac data set. "Ab." and "Ca." are abbreviations for Abdominal and Cardiac. "Best" refers to the best performance of the model in 200 epochs. "Entropy" contains three states: "-" not used; "Min" entropy minimization; "Max" entropy maximization. }
 \begin{center}
 \scalebox{0.8}{
  \begin{tabular}{c|ccccc|ccccc}
    \toprule
    Group & Ab. $CT \rightarrow MRI$ & Entropy & Epoch 50 & Epoch 200 &Best & Ab. $MRI \rightarrow CT$ & Entropy & Epoch 50 & Epoch 200 &Best \\
    \midrule
    \multirow{3}*{A}&fairLD  & - &0.5823&0.5312&0.6832&fairLD&-&0.6772&0.6219&0.7101 \\
    ~&OS &-&0.6772&0.6751&0.6878&OS&-&0.6009&0.5722&0.6611 \\
    ~&DPL &-&0.5375&0.4912&0.6692&DPL&-&0.6097&0.5792&0.6534 \\
    \hline
    \multirow{3}*{B}&fairLD & Min &0.5578&0.5053&0.6861&fairLD&Min&0.6532&0.5979& 0.7061 \\
    ~&OS & Min&0.6765&0.6698&0.6846&OS&Min&0.5835&0.5301&0.6580 \\
    ~&DPL &Min&0.5198&0.4789&0.6718&DPL&Min&0.5977&0.5521&0.6598 \\

    \hline
    \multirow{3}*{C}&fairLD+WC & - &0.6811&0.6679&0.7011&fairLD+WC&-&0.7098&0.6881&0.7159  \\
    ~&OS+WC & -&0.6771&0.6698&0.6834&OS+WC&-&0.6375&0.6198&0.6639 \\
    ~&DPL+WC &-&0.6858&0.6699&0.6912&DPL+WC&-&0.6462&0.6219&0.6559 \\
    \hline
    \multirow{3}*{D}&fairLD+WC & Max &0.6821&0.6897&0.7006&fairLD+WC&Max&0.7128&0.7072&0.7188  \\
    ~&OS+WC & Max&0.6673&0.6733&0.6867&OS+WC&Max&0.6304&0.6408&0.6606 \\
    ~&DPL+WC &Max&0.6866&0.6881&0.6961&DPL+WC&Max&0.6426&0.6482& 0.6535\\
    \bottomrule
  \end{tabular}
}
 \scalebox{0.8}{
  \begin{tabular}{c|ccccc|ccccc}
    \toprule
    Group & Ca. $CT \rightarrow MRI$ & Entropy & Epoch 50 & Epoch 200 &Best & Ca. $MRI \rightarrow CT$ & Entropy & Epoch 50 & Epoch 200 &Best \\
    \midrule
    \multirow{3}*{A}&fairLD  & - &0.3823&0.3629&0.4233&fairLD&-&0.6021&0.5641& 0.6254\\
    ~&OS &-&0.4177&0.4152&0.4234&OS&-&0.4477&0.4202&0.4596 \\
    ~&DPL &-&0.3961&0.3638&0.4238&DPL&-&0.5049&0.4593&0.5447 \\
    \hline
    \multirow{3}*{B}&fairLD & Min &0.3661&0.3329&0.4202&fairLD&Min&0.5829&0.5545&0.6234  \\
    ~&OS & Min&0.4186&0.4159&0.4261&OS&Min&0.4375&0.4149&0.4609 \\
    ~&DPL &Min&0.3872&0.3511&0.4244&DPL&Min&0.4982&0.4521&0.5455 \\

    \hline
    \multirow{3}*{C}&fairLD+WC & - &0.4144&0.4003&0.4283&fairLD+WC&-&0.6218&0.6022&0.6247  \\
    ~&OS+WC & -&0.4102&0.3998&0.4232&OS+WC&-&0.4899&0.4728&0.5052 \\
    ~&DPL+WC &-&0.4254&0.3978&0.4225&DPL+WC&-&0.5421&0.5372&0.5583 \\
    \hline
    \multirow{3}*{D}&fairLD+WC & Max &0.4136&0.4182&0.4264&fairLD+WC&Max&0.6223&0.6198&0.6289  \\
    ~&OS+WC & Max&0.4141&0.4164&0.4202&OS+WC&Max&0.4912&0.4955&0.5083 \\
    ~&DPL+WC &Max&0.4213&0.4129&0.4238&DPL+WC&Max&0.5456&0.5508&0.5608 \\
    \bottomrule
  \end{tabular}
  }
 
  \label{tab:ablation}
  \end{center}

\end{table*}

\textbf{The role of self-entropy minimization in SFUDA tasks.} The experiments of group A and group B in the Table \ref{tab:ablation} are related to this problem. The experimental results in group A did not use entropy minimization strategy and the experimental results in group B used it. From the best performance of the two groups, it can be found that entropy minimization strategy can not improve the best performance of SFUDA methods. For example, in abdominal $CT \rightarrow MRI$ task, the best performance of fairLD, OS and DPL without entropy minimization are 0.6832, 0.6878 and 0.6692 respectively. They did not improve significantly after using entropy minimization (0.6861, 0.6846 and 0.6718). However, entropy minimization might aggravate the over-fitting phenomenon. The experimental results show that the performance of the method using entropy minimization at 50th epoch50 and 200th epoch are worse than those without entropy minimization. In general, the effect of self-entropy minimization in SFUDA tasks can not play a significant positive impact.

\textbf{The role of Stable Learning.} The experiments of group D in the Table \ref{tab:ablation} are related to this problem. For the effectiveness and stability of ST in Table \ref{tab:abnorminal}, we have considered such a possibility: by slowing down the convergence speed of the model, the process of self-training is stabilized. If this assumption holds, it means that if the training epoch is prolonged, the final model will still fall into the dilemma of over-fitting. We extended the training epoch from 50 to 200, and found that ST can indeed stabilize the training process, not delay the convergence speed.  In Group D, the performance of fairLD with SL changed from 0.6821 (50th epoch) to 0.6897 (200th epoch) in abdominal $CT \rightarrow MRI$ task. Similarly, the performance of fairLD with SL changed from 0.6223 (50th epoch) to 0.6198 (200th epoch) in cardiac $MRI \rightarrow CT$ task. In general, the effect of Stable Learning can indeed stabilize the model and avoid the model falling into over-fitting dilimme.

\textbf{The role of entropy increase in stable learning.} The experiments of group C and group D in the Table \ref{tab:ablation} are related to this problem. The experimental results in group C did not use entropy increase strategy and the experimental results in group D used it. Weight Consolidation was used in both groups. From the "Epoch 200" of the two groups, it can be found that the method with WC strategy only had a decline in performance after a long time of training. For example, in abdominal $CT \rightarrow MRI$ task, the performance of fariLD (Group C) suffer a decline from 0.6811 (Epoch 50) to 0.6679 (Epoch 200); that of DPL also has a drop from 0.6858 (Epoch 50) to 0.6699 (Epoch 200). However, the performance of fariLD and DPL (Group D) is more stable, which is from 0.6821, 0.6866 (Epoch 50) to 0.6897, 0.6881 (Epoch 200) respectively. The reason for this phenomenon is that overparameterization makes it possible to have a solution to make the model suffer the same over-fitting dilemma under the constraint of weight consolidation (more details can be found in section \ref{sec:EI}).

\section{Conclusion}
We introduce a Stable Learning (ST) strategy in source-free domain adaption for medical semantic segmentation. We point out that the existing SFUDA methods suffer an over-fitting dilemma without a labeled target-domain validation data set. The proposed Stable Learning contains two components: Weight Consolidation and Entropy Increase. SL strategy is compatible with multiple existing medical SFUDA methods, and the effectiveness has been verified in four SFUDA tasks. In addition, our upcoming release of the SFUDA library will make an active contribution to future research.


%
\bibliographystyle{splncs04}

\bibliography{./AP.bib}
%

\end{document}